\documentclass[letterpaper]{article} 
\usepackage{aaai23}  
\usepackage{times}  
\usepackage{helvet}  
\usepackage{courier}  
\usepackage[hyphens]{url}  
\usepackage{graphicx} 
\usepackage{natbib}  
\usepackage{caption} 
\usepackage{graphicx}
\usepackage{amsmath}
\usepackage{amssymb}
\usepackage{amsthm}
\usepackage{thmtools}
\usepackage{mathtools}
\usepackage{subfig}

\usepackage{xcolor}

\usepackage[protrusion=true,expansion=true]{microtype}	
\allowdisplaybreaks
\newtheorem{definition}{Definition}

\newtheorem{lemma}{Lemma}

\usepackage[toc,page]{appendix}
\newcommand{\ignore}[1]{} 

\usepackage{algorithm}
\usepackage{algorithmic}

\usepackage{newfloat}
\usepackage{listings}
\DeclareCaptionStyle{ruled}{labelfont=normalfont,labelsep=colon,strut=off} 
\floatstyle{ruled}
\newfloat{listing}{tb}{lst}{}
\floatname{listing}{Listing}
\title{Provable Detection of Propagating Sampling Bias in Prediction Models}
\author{
    Pavan Ravishankar \textsuperscript{\rm 1},
    Qingyu Mo \textsuperscript{\rm 1},
    Edward McFowland III \textsuperscript{\rm 2},
    Daniel B. Neill \textsuperscript{\rm 1}
}
\affiliations{
    \textsuperscript{\rm 1}Machine Learning for Good Laboratory, New York University;
     \textsuperscript{\rm 2}Harvard Business School\\

    pr2248@nyu.edu, qm348@nyu.edu, emcfowland@hbs.edu, daniel.neill@nyu.edu
}

\usepackage{bibentry}

\begin{document}

\maketitle

\begin{abstract}
With an increased focus on incorporating fairness in machine learning models, it becomes imperative not only to assess and mitigate bias at each stage of the machine learning pipeline but also to understand the downstream impacts of bias across stages. Here we consider a general, but realistic, scenario in which a predictive model is learned from (potentially biased) training data, and model predictions are assessed post-hoc for fairness by some auditing method.  We provide a theoretical analysis of how a specific form of data bias, differential sampling bias, propagates from the data stage to the prediction stage. Unlike prior work, we evaluate the downstream impacts of data biases quantitatively rather than qualitatively and prove theoretical guarantees for detection.  Under reasonable assumptions, we quantify how the amount of bias in the model predictions varies as a function of the amount of differential sampling bias in the data, and at what point this bias becomes provably detectable by the auditor.  Through experiments on two criminal justice datasets-- the well-known COMPAS dataset and historical data from NYPD's stop and frisk policy--  we demonstrate that the theoretical results hold in practice even when our assumptions are relaxed.
\end{abstract}

\section{Introduction}
Machine learning models are being used in numerous applications such as healthcare \cite{de2018clinically}, online advertising \cite{perlich2014machine}, and finance \cite{malekipirbazari2015risk}. Due to its increased proliferation, there is a rising concern in the machine learning community to deploy fair machine learning models \cite{barocas2017fairness,mehrabi2021survey}. Since decision-making in machine learning comprises of various stages such as the data stage, modeling stage, and prediction stage \cite{suresh2019framework}, it becomes imperative to look at the fairness problem across stages, rather than limiting the discussion to a single stage. For instance, the \textbf{data stage} could be biased due to members of a subgroup being systematically selected with a higher or a lower probability than others \cite{samplingbias}, also known as sample selection bias. Such biases could propagate to the \textbf{prediction stage}, and the resulting biases in prediction could be compounded by other sources such as model misspecification~\cite{gajane2017formalizing}.  However, it is unclear precisely how and to what extent the data bias would affect the predictions, and when the resulting prediction biases would be detectable by some auditing approach.  Such biases, once detected and precisely characterized, could then be corrected, e.g., by resampling to de-bias the data.

In this paper, we analyze the propagation of \textbf{differential sampling bias} from the data stage to the prediction stage.  Differential sampling bias is a form of sample selection bias in which some subpopulation $S$ is sampled non-uniformly, such that the distribution of an outcome variable $Y$ given predictor variables $X$ in the sampled data for $S$ differs from the true (population) distribution of $Y$ given $X$ for $S$.  \footnote{Note that differential sampling bias would \emph{not} be present if subpopulation $S$ was under- or over-sampled but the distribution of $Y$ given $X$ for $S$ remained unchanged.  We do not address other forms of sample selection bias here.} 

This bias can arise in many different circumstances. For example, in criminal justice, both the organizational biases of police departments (e.g., a policy of conducting large numbers of pedestrian stops in predominantly minority neighborhoods) and the perceptual biases of individual police officers (e.g., higher likelihood of stopping and frisking Black individuals) led to much higher proportions of Black individuals being arrested for marijuana possession, despite similar rates of use in the population as a whole~\cite{aclu2020}. In our analysis of NYPD stop and frisk data, we consider the race of the stopped individual as our outcome variable, and observe that Pr(race = ``Black'') is significantly increased as compared to a ``less biased'' alternative policing strategy.  \ignore{\footnote{If we had instead focused on the discovery of contraband as our outcome variable, we would observe a \emph{decreased} probability for Black individuals, \emph{conditional on being stopped}, since stops were made with a lower degree of suspicion by police.}} 

Differential sampling bias can also result from \textbf{concept shift}: a model meant for prediction of outcome variable $Y$ in one setting is learned using data from a different setting where the relationship between $Y$ and the predictor variables $X$ differs.  For example, if criminal justice data from one jurisdiction is used to predict a defendant's risk of reoffending in a different jurisdiction, or if historical data is used and reoffending patterns have changed over time, the training data will exhibit differential sampling bias: the proportion of reoffenders for certain demographics may be higher or lower in the training data as compared to the true probabilities for the jurisdiction and time period of interest.  In our experimental analysis of the COMPAS dataset, we inject simulated differential sampling bias (assuming concept shift) by weighted resampling of the training data.
\clearpage
\begin{figure}[t]
    \centering
        \includegraphics[width=0.9\columnwidth, height=0.1\columnwidth]{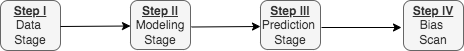}
        \caption{Problem setup. \textbf{Step 1:} Differential sampling bias $\Delta$ is induced into subgroup $S^T$. \textbf{Step 2:} A classification model is trained on the biased data to predict the probability of belonging to class $Y=1$. \textbf{Step 3:} Predicted probabilities of belonging to $Y=1$, given by $\{\widetilde{p}_i\}$, are made on the test data. \textbf{Step 4:} Bias Scan finds the most biased subgroup $S^\ast$ and its log-likelihood ratio score $F^\ast$ based on the predictions.} \label{setup}
\end{figure}

Here we introduce the first formal analysis of how differential sampling bias induced in the data stage (i.e., biased training data) propagates through the modeling and prediction stages, leading to significant biases in prediction. These propagated biases can then be detected by an \emph{auditor} that compares the model predictions with the observed outcomes.  

Our \textbf{problem setup} is shown in Figure~\ref{setup}: first, in the \emph{data stage}, we assume initially unbiased training and test data records drawn i.i.d.~from some joint probability distribution $f_{X,Y}(x,y)$ of the predictor variables $X$ and binary outcome variable $Y$.  Then differential sampling bias $\Delta$ is injected into the ``true'' subgroup $S^T$ for the training data only.  Without loss of generality, we define $Y$ such that the differential sampling bias increases the probability $\widetilde{\mathbb{P}} (Y=1|X)$, thus over-sampling records with $Y=1$ in subgroup $S^T$. We parameterize the multiplicative increase in the odds of $Y=1$ by $\Delta > 1$.  Second, in the \emph{modeling stage}, a classification model is trained using the biased data. Third, in the \emph{prediction stage}, the classifier makes predictions $\widetilde p_i$ (the estimated probability that $Y=1$ for each data record) for the test data. Fourth, Bias Scan~\cite{biasscan} is used to assess whether the predictions $\widetilde p_i$ are systematically biased as compared to the test outcomes $y_i$ for any intersectional subgroup.

Given this problem setup, we present theoretical and empirical results showing (a) the amount of bias that propagates from the data stage to the prediction stage, as measured by the log-likelihood ratio (LLR) score found by Bias Scan; and (b) when the bias will exceed a threshold for significance, assuming a fixed false positive rate $\alpha$, thus enabling detection by Bias Scan.  Our specific \textbf{contributions} are as follows:
\begin{enumerate}
\item We define and quantify the differential sampling bias $\Delta$ induced into subgroup $S^T$ in the binary outcome $Y$.
\item We derive a new closed-form expression for the LLR score of Bias Scan, used to audit a consistent classifier trained on large data with differential sampling bias.
\item We present a new asymptotic result for the null distribution of the Bias Scan score, which leads to a threshold score $h(\alpha)$ for detection at a fixed false positive rate $\alpha$.
\item We demonstrate detection with full asymptotic power, $\mathbb{P}_{H_1}$(Reject $H_0$) $\rightarrow 1$, as the data size becomes large.
\item Using the threshold $h(\alpha)$, we find the minimum amount of bias $\Delta$ that needs to be induced in subgroup $S^T$ for it to be provably detectable in the finite sample case.
\item We evaluate our theoretical results empirically on two different criminal justice datasets.  On the well-known COMPAS dataset, we compare the empirical and theoretical relationships between the Bias Scan score $F^\ast$ and the amount of injected bias $\Delta$, across two different classification models and two types of bias injection (marginal and intersectional). We also analyze historical data from the NYPD's ``stop-question-frisk'' (SQF) policy, estimating the amount of differential sampling bias $\Delta$ in the data as compared to a ``less biased'' alternative policing strategy.
\item For both datasets, we observe that the empirical relationship between the propagated bias in predictions (as measured by the Bias Scan score $F^\ast$) and the differential sampling bias in data (as measured by $\Delta$) corresponds well to the theoretical values. We also confirm that, if enough bias is present in the data stage, then the affected subgroup is detectable by the auditor in the prediction stage with high accuracy. These two conclusions demonstrate the validity of the theoretical assumptions and provide reasoning when theoretical and empirical results differ.
\end{enumerate}

\section{Related Work}
\textbf{Stage-specific notions of fairness and bias:} The machine learning community has typically centered the fairness problem in either the data stage or the prediction stage \cite{barocas2017fairness}. In the data stage, various attempts have been made to detect and mitigate data biases.  For example, \citet{zemel2013learning}, \citet{madras2018learning}, and \citet{song2019learning} attempt to de-bias data by learning fair representations. \citet{silvia2020general}, \citet{oneto2020fairness}, and \citet{ravishankar2021causal} discuss causal notions of fairness such as path-specific fairness, and use them to detect and mitigate unfairness in the data generation process. Similarly, many approaches have been proposed to address biases in the prediction stage: 
\citet{berk2021fairness} state multiple fairness definitions such as demographic parity and calibration based on model predictions; \citet{corbett2018measure} discuss the limitations of these fairness definitions; \citet{kleinberg2016inherent} and \citet{chouldechova2017fair} prove that, except in special cases, these definitions are incompatible; \citet{zadrozny2004learning} proposes a framework to correct bias in model predictions; and \citet{pedreschi2009measuring} propose novel measures of discrimination to correct discriminatory patterns. None of the aforementioned works have analyzed how bias \textit{propagates} downstream, across different stages of the pipeline.

\textbf{Bias propagation pipelines:} \citet{suresh2019framework} discuss the bias problem holistically, rather than centering it to a particular stage, by laying out a framework comprising of biases originating at different stages of the pipeline. Similarly, an opinion article by \citet{hooker2021moving} proposes that bias should be viewed and analyzed as an aggregation of the biases arising in different stages.  However, neither of these works provide any formal, quantitative analysis of how bias propagates between stages. \citet{rambachan2019bias} quantitatively analyze how selection bias propagates from the data stage to the prediction stage. However, the study makes a strong assumption about the form of the selection process, and does not discuss whether the propagated bias is detectable or how it can be detected in the prediction stage.

\textbf{Frameworks for detection of intersectional biases:}  Several recent approaches have been proposed to detect biases affecting a subpopulation defined along multiple data dimensions ~\cite{biasscan,kearns18a}. Here we apply Bias Scan~\cite{biasscan} to assess models learned from biased data, detecting intersectional subgroups where the model predictions $\widetilde p_i$ most significantly overestimate $\mathbb{P}(Y = 1 \:|\: X=x_i)$.
Bias Scan builds on previous univariate and multivariate subset scan approaches~\cite{neill2012fast,neill2013fast}. Additionally, \citet{mcfowland2018efficient} use a similar multidimensional scan framework to discover the subgroups that are most significantly affected by a treatment in a randomized experiment, and provide statistical guarantees on detection.
However, all of these approaches focus on a single pipeline stage (predictions or outcomes), while our work examines the propagation of data biases into model predictions.

\section{Preliminaries}
\subsection{Notations}
Assume test data $D = \{(x_i,y_i)\}$ drawn i.i.d.~from joint probability distribution $f_{X,Y}(x,y) = f_X(x) f_{Y|X}(y|x)$ and training data
$\widetilde D = \{(\widetilde x_i,\widetilde y_i)\}$ drawn i.i.d.~from joint probability distribution $\widetilde f_{X,Y}(x,y) = \widetilde f_X(x) \widetilde f_{Y|X}(y|x)$. Here $Y$ is a binary outcome variable, and thus we can write $f_{Y|X}(y|x)$ and
$\widetilde f_{Y|X}(y|x)$ as the
probabilities $\mathbb{P}(Y=y\:|\:X=x)$ and $\widetilde{\mathbb{P}}(Y=y\:|\:X=x)$, $y \in \{0,1\}$,
for test and training data respectively.  Let $p_i = \mathbb{P}(Y=1\:|\:X=x_i)$ be the true probability that $Y=1$ for test record $s_i = (x_i, y_i)$, and let $\widetilde p_i$ and $\hat p_i$
be the estimated probabilities that $Y=1$ for test record $s_i$
from classification models learned from training data with and without differential sampling bias. Note that $\widetilde p_i = \hat p_i$ when $\Delta=1$ (under the null hypothesis of no bias).

We assume that $X$ consists of a set of discrete-valued\footnote{Sensitive covariates (e.g. race, ethnicity, and gender) are usually discrete. Continuous covariates can be discretized as a preprocessing step, using the observed covariate distribution or domain knowledge.} predictor variables $\{ X_1, \ldots, X_Q \}$ and that each variable $X_i$ takes on a set of values $V_i$.  An intersectional \textbf{subgroup} $S$ is defined as a subset of the Cartesian product $V = V_1 \times \ldots \times V_Q$. A \textbf{rectangular} subgroup is one that can be represented as the Cartesian product of subsets of attribute values, $S = S_1 \times \ldots \times S_Q$, for $S_i \subseteq V_i$.  For example, if $X_1$ = Gender, $V_1$ = \{Male, Female\}, $X_2$ = Race, and $V_2$ = \{Black, White, Other\}, then \{Male, Female\} $\times$ \{Black, White\} = \{(Male, Black), (Female, Black), (Male, White), (Female, White)\} is a rectangular subgroup, while \{(Male, Black), (Female, White)\} is non-rectangular. Let $rect(X)$ denote the set of all rectangular subgroups of $X$. Finally, for test dataset $D$, we associate with any given subgroup $S$ the subset of matching data records $D_S = \{(x_i,y_i)\} \subseteq D: x_i \in S$.

\subsection{Bias Scan}
Bias Scan~\cite{biasscan} is a multi-dimensional subset scanning algorithm used to detect intersectional subgroups for which a classifier's probabilistic predictions $\widetilde p_i$ of a binary outcome $y_i$ are significantly biased as compared to the observed outcomes $y_i$.  More precisely, Bias Scan searches for the rectangular subgroup $S^\ast$ which maximizes a Bernoulli log-likelihood ratio (LLR) scan statistic,\footnote{We assume that the bias is injected into a rectangular subgroup, a common formulation (e.g., used in decision trees), as it is representative of a cohesive and interpretable subpopulation.}
\[S^\ast = \underset{S \in rect(X)}{\arg\max} ~F(S),\] with corresponding LLR score $F^\ast = F(S^\ast)$.

To obtain the score function for a given subgroup $S$, Bias Scan computes the generalized log-likelihood ratio
$F(S) = \underset{\widetilde q}{\max}\log \frac{P(D \:|\: H_1(S,\widetilde q))}{P(D \:|\: H_0)}$, assuming the following hypotheses:
\begin{align*}
    &H_0 : &\text{odds}(y_i) = \frac{\widetilde{p}_i}{1-\widetilde{p}_i}, \:\: &\forall s_i \in D.~~~~~~~~ \\
    &H_1(S,\widetilde q) : &\text{odds}(y_i) = \frac{\widetilde{q} \: \widetilde{p}_i} {1-\widetilde{p}_i}, \:\: &\forall s_i \in D_S, \\ & &\text{odds}(y_i) = \frac{\widetilde{p}_i}{1-\widetilde{p}_i}, \:\: &\forall s_i \in D \setminus D_S.
\end{align*}

Here we detect biases where the probabilities $\widetilde p_i$ are over-estimated, and thus $0 < \widetilde q < 1$. As derived in the Technical Appendix, the resulting log-likelihood ratio score $F(S)$ is
\begin{equation}
    F(S) = \max_{0 < q < 1} \left( \sum_{s_i \in D_S} y_i \log q - \sum_{s_i \in D_S} \log (1 - p_i + q\:p_i) \right). 
    \label{bias_scan_score}
\end{equation}

The Bias Scan algorithm for optimizing $F(S)$ over rectangular subgroups is provided in the Technical Appendix.

\section{Differential Sampling Bias}
In this section, we quantify differential sampling bias for a subgroup $S$, as follows:

\begin{definition}
A subgroup $S$ exhibits \textbf{differential sampling bias} $\Delta > 1$ towards the outcome $Y=1$ if, for all $x \in S$,
\begin{align}
    \widetilde{\mathbb{P}}(Y=1|X=x) = \frac{\Delta\mathbb{P}(Y=1|X=x)}{\Delta\mathbb{P}(Y=1|X=x) + \mathbb{P}(Y=0|X=x)}.\label{diffbiasdefn}
\end{align} 
\end{definition}

For example, differential sampling bias could be injected into unbiased training data by re-drawing data elements $\{(\widetilde x_i,\widetilde y_i)\}$, for $\widetilde x_i \in S$, with replacement, with sampling weights $w_i = \Delta$ for $\widetilde y_i = 1$ and $w_i = 1$ for $\widetilde y_i = 0$. We use this approach to inject bias into the COMPAS dataset in our experiments below. \ignore{In this case, we note that the probability $\mathbb{P}(X \in S)$ remains constant \textbf{(DOUBT 2)}. Moreover, for $X \in S$,
 $\widetilde f_{X|Y} = f_{X|Y}$, while for $X \not\in S$, $\widetilde f_{X,Y} = f_{X,Y}$. 
However, these additional properties are not needed for the theoretical results below, and do not hold for our experiments using the NYPD stop-question-frisk dataset.}

\ignore{
\subsection{Notations}

\noindent 
$U_X(S)$ is the set of unique co-variate profiles in $S$.\\
$N(x_i)$ is the number of test points with $Y=1$ in profile $x_i$.\\
$N_{S,Y=1}$ is the number of test points in $S$ with $Y=1$.\\
$S^{*}_{\text{u}}$ is the unconstrained subgroup that maximizes $F(S)$.\\
$M$ is the number of unique co-variate profiles.\\
$M_i$ is the number of unique profiles with probability $p_i$ for $Y=1$.\\
$n$ is the average number of test data with the same co-variate profile.\\
$v$ is the fraction of test data contained in the sub-group to which bias is injected. 
}

\section{Theoretical Results}
In this section, we derive theoretical results to understand the propagated effects of differential sampling bias and to provide statistical guarantees for detectability.

More precisely, we prove four main theorems.  Given the problem setup described above and the assumptions listed below, \textbf{Theorem~\ref{thm:thm1}} provides an asymptotic closed-form formulation of the Bias Scan log-likelihood ratio (LLR) score $F(S^T)$ of the injected subgroup $S^T$ as a function of the amount of differential sampling bias $\Delta$. If $S^T$ is a rectangular subgroup, $S^T \in rect(X)$, this score is a lower bound on the overall Bias Scan score $F^\ast = \max_{S\in rect(X)} F(S)$.  \textbf{Theorem~\ref{thm:thm2}} provides an upper bound for the null distribution of $F^\ast$ (i.e., assuming no bias is present), enabling us to compute a threshold score for detection.  Finally, \textbf{Theorems~\ref{thm:thm3} and~\ref{thm:thm4}} combine these results to show asymptotic detection with full power for any $\Delta > 1$ as the sizes of the training and test data go to infinity, as well as computing the minimum amount of bias $\Delta$ needed for detection in finite test data.

These Theorems rely on \textbf{three key assumptions}:
\begin{enumerate}
    \item[(A1)] \emph{Consistency} of the classifier used in the prediction stage, for learning the conditional distribution $\widetilde f_{Y|X}$.
    \item[(A2)] \emph{Full support} of the biased training data: $support(\widetilde f_X) \supseteq support(f_X)$, and $support(f_X) \cap S \ne \emptyset$. 
    \item[(A3)] \emph{Positivity}: $0 < \mathbb{P}(Y=1 \:|\: X=x)  < 1, \: \forall x$.
\end{enumerate}

Given these assumptions, we first derive the relationship between the amount of differential sampling bias $\Delta$ injected into subgroup $S$, and the Bias Scan score $F(S)$:

\begin{restatable}{theorem}{firstthm}
Assume that a classifier is trained on data $\widetilde D$ with differential sampling bias $\Delta > 1$ for subgroup $S$ and makes predictions $\widetilde p_i$ for unbiased test data $D = \{(x_i,y_i)\}$.  If Bias Scan is used to assess bias in $\widetilde p_i$ as compared to $y_i$, then under assumptions (A1)-(A3), as the number of training data records $|\widetilde D| \rightarrow \infty$, the Bias Scan score $F(S)$ of subgroup $S$ converges to:
\[
F(S) \rightarrow 
F_{old}(S) - \sum_{s_i \in D_{S}} y_i \log \Delta + \sum_{s_i \in D_{S}} \log (\Delta p_i + 1 - p_i),
\]
if $\Delta > \hat q_{MLE}$, and $F(S) \rightarrow 0$ otherwise, where $\hat q_{MLE}$ is the maximum likelihood estimate of $\widetilde q$ for Bias Scan assuming no differential sampling bias ($\Delta = 1)$, satisfying
\[\sum_{s_i \in D_S} y_i = \sum_{s_i \in D_S}
\frac{\hat q_{MLE}\:p_i}{\hat q_{MLE}\:p_i + 1 - p_i},
\]
and
\[F_{old}(S) = \sum_{s_i \in D_S} y_i \log \hat q_{MLE} - \sum_{s_i \in D_S} \log(1-p_i+\hat q_{MLE}\:p_i)\]
is the Bias Scan score of subgroup $S$ assuming no differential sampling bias ($\Delta = 1$).
\label{thm:thm1}
\end{restatable}

The proof of Theorem~\ref{thm:thm1} is provided in the Technical Appendix.  Critically, under assumptions (A1)-(A3), as $|\widetilde D| \rightarrow \infty$, we have  $\widetilde p_i \rightarrow \widetilde {\mathbb{P}}(Y=1 \:|\: X=x_i) = \frac{\Delta p_i}{\Delta p_i + 1 - p_i}$ for all $s_i \in D_S$, and the corresponding predicted probabilities with no differential sampling bias, $\hat p_i \rightarrow {\mathbb{P}}(Y=1 \:|\: X=x_i)  = p_i$.  We then show that the maximum likelihood estimate (MLE) of $\widetilde q$ for Bias Scan is $\hat q_{MLE} / \Delta$, where $\hat q_{MLE}$ is the corresponding MLE with no differential sampling bias.  Finally, we plug in the expressions for $\widetilde p_i$, $\hat p_i$, and $\widetilde q_{MLE}$, and simplify.

\begin{restatable}{corollary}{firstcorr}
Under the conditions of Theorem~\ref{thm:thm1}, as the number of test data records $|D| \rightarrow \infty$, the normalized Bias Scan score $F(S)/|D|$ of subgroup $S$ converges to:
\[\frac{F(S)}{|D|} \rightarrow \mathbb{P}(x \in S) \mathbb{E}_{s_i \in D_{S}} [\log (\Delta p_i + 1 - p_i) - p_i \log \Delta], \]
an increasing function of $\Delta$.
\label{corr:corr1}\end{restatable}

Next, we provide statistical guarantees for the detection of bias.  To do so, we first consider the distribution of the Bias Scan score $F^\ast = \max_{S \in rect(X)} F(S)$ under the null hypothesis of no bias, $H_0$.  For a given false positive rate $\alpha$, we find a score threshold $h(\alpha)$ such that $\mathbb{P}_{H_0}(F^\ast > h(\alpha)) \le \alpha$.  

To do so, we make the additional assumption:
\begin{enumerate}
    \item[(A4)] The number of unique covariate profiles in the test data, $M$, is large enough so that Gaussian approximations hold (e.g., $M>30$) but finite (i.e., $M$ remains constant as the number of test data records $|D| \rightarrow \infty$).
\end{enumerate}

Then we can show the following:
 
\begin{restatable}{theorem}{secondthm}
Assume that a classifier is trained on unbiased training data $\widetilde D$ and makes predictions $\hat p_i$ for unbiased test data $D = \{(x_i,y_i)\}$, and Bias Scan is used to assess bias in $\hat p_i$ as compared to $y_i$.
Let $F^\ast = \max_{S \in rect(X)} F(S)$ be the Bias Scan score, maximized over all rectangular subgroups $S$.  Then under assumptions (A1)-(A4), as the number of training data records $|\widetilde D| \rightarrow \infty$ and the number of test data records $|D| \rightarrow \infty$,
for a given Type-I error rate $\alpha > 0$,
there exists a critical value $h(\alpha)$ and constants $k_1 \approx 0.202$, $k_2 \approx 0.523$ such that
\[
    \mathbb{P}(F^\ast > h(\alpha)) \leq \alpha,
\]
where
\begin{align}
    h(\alpha) = k_1 M + k_2 \Phi^{-1}(1-\alpha) \sqrt{M},
    \label{eqn:halpha}
\end{align}
and $\Phi$ is the Gaussian cdf.
\label{thm:thm2}\end{restatable}

Critically, $h(\alpha)$ does not depend on the number of test data records $|D|$, but only on the number of unique covariate profiles in the test data $M$.  Now, we prove that under the presence of bias $\Delta$, $h(\alpha)$ serves as a threshold for rejecting the null hypothesis of no bias with full asymptotic power.

\begin{restatable}{theorem}{thirdthm}
Assume that a classifier is trained on data $\widetilde D$ with differential sampling bias $\Delta > 1$ for rectangular subgroup $S^T$ and makes predictions $\widetilde p_i$ for unbiased test data $D = \{(x_i,y_i)\}$, and Bias Scan is used to assess bias in $\widetilde p_i$ as compared to $y_i$. Let $F^\ast = \max_{S \in rect(X)} F(S)$ be the Bias Scan score, and let $h(\alpha)$ be the score threshold for detection at a fixed Type-I error rate of $\alpha$, as given in Equation (\ref{eqn:halpha}).  Then for any $\alpha > 0$ and $\Delta > 1$, under assumptions (A1)-(A4), as the number of training data records $|\widetilde D| \rightarrow \infty$ and the number of test data records $|D| \rightarrow \infty$, $\mathbb{P}(F^\ast > h(\alpha)) \rightarrow 1$.
\label{thm:thm3}\end{restatable}

We now find the minimum bias that needs to be induced into subgroup $S$ to be detectable for a given Type-I error rate.
\begin{restatable}{theorem}{fourththm}
Assume that a classifier is trained on data $\widetilde D$ with differential sampling bias $\Delta > 1$ for rectangular subgroup $S^T$ and makes predictions $\widetilde p_i$ for unbiased test data $D = \{(x_i,y_i)\}$, and Bias Scan is used to assess bias in $\widetilde p_i$ as compared to $y_i$. Let $F^\ast = \max_{S \in rect(X)} F(S)$ be the Bias Scan score, and let $h(\alpha)$ be the score threshold for detection at a fixed Type-I error rate of $\alpha$, as given in Equation (\ref{eqn:halpha}).  Further, assume $D_{S^T}$ is fixed, with finite size $|D_{S^T}|$ and $\left(\sum_{s_i \in D_{S^T}} y_i \right) < |D_{S^T}|.$  Then for any $\alpha > 0$,  under assumptions (A1)-(A4), as the number of training data records $|\widetilde D| \rightarrow \infty$, there exists $\Delta_{thresh} \ge 1$ such that, if $\Delta > \Delta_{thresh}$, then $\mathbb{P}(F^\ast > h(\alpha)) \rightarrow 1$, where
\[ \Delta_{thresh} = \max(1,Q^{-1}(h(\alpha) - F_{old}(S^T))),\]
\[Q(\Delta) = \sum_{s_i \in D_{S^T}} (\log (\Delta p_i + 1 - p_i) - y_i \log \Delta),\]
and $F_{old}(S^T)$ is the Bias Scan score of subgroup $S^T$ assuming no differential sampling bias ($\Delta = 1$).
\label{thm:thm4}\end{restatable}
Proofs of Theorems \ref{thm:thm1}-\ref{thm:thm4} are provided in the Appendix.

\section{Experiments}
We perform experiments on two criminal justice datasets to validate our theoretical results: semi-synthetic predictions of recidivism risk derived from the well-known COMPAS dataset, and real-world ``stop, question and frisk'' (SQF) data from the New York Police Department (NYPD). 

\subsection{Experiments on COMPAS/ProPublica Data}
COMPAS is a commercial decision-support algorithm which has been applied in many jurisdictions to estimate a defendant's probability of reoffending, with impacts on criminal justice outcomes such as bail, sentencing, and parole. COMPAS gained notoriety when investigative journalists from ProPublica published a study arguing that COMPAS was racially biased against Black defendants~\cite{angwin2016machine}. The public dataset compiled by ProPublica\footnote{https://github.com/propublica/compas-analysis/compas-scores-two-years.csv}, including COMPAS risk predictions for 7,214 defendants in Broward County, Florida, from 2013-2014, and a two-year follow-up to record which defendants were rearrested, has been studied by numerous algorithmic bias researchers~\cite{barenstein2019propublica}.  

While most of these analyses focus on assessing biases in the COMPAS risk predictions~\cite{chouldechova2017fair,kleinberg2018algorithmic}, we instead utilize this dataset to learn predictive models for the binary outcome (rearrest within two years) as a function of five categorical predictor variables\footnote{Predictors include gender, race, charge degree, $\text{age}<25$, and number of prior offenses (``none'', ``1 to 5'', or ``more than 5'').}, and use these models to study how differential sampling bias in the data propagates to the model predictions.

To do so, we consider differential sampling biases $\Delta \in \{1, 1.25, 1.5, \ldots, 10\}$ injected into one of two rectangular subgroups. Letting $X_1$ = Gender, $X_2$ = Race, and $V_j$ = the set of all possible values for attribute $X_j$, we consider the subgroups $S^T = \{\text{Female}\} \times V_2 \times \ldots \times V_5$ and $S^T = \{\text{Female}\} \times \{\text{Caucasian}\} \times V_3 \times \ldots \times V_5$.  The first subgroup represents a \emph{marginal bias} against females (since we are oversampling females who reoffended, as compared to females who did not reoffend, by a factor of $\Delta$ in the training data, thus leading to an overestimate of their reoffending risk), while the second subgroup represents an \emph{intersectional bias} against white females.  We also consider two different classifiers, random forest and logistic regression, and average results over 100 trials for each combination of classifier, injected subgroup $S^T$, and amount of bias $\Delta$.
\ignore{
We conduct experiments using different classifiers: the random forest classifier, logistic regression classifier, and ideal predictor\footnote{Ideal Predictor shifts COMPAS predictions for the subset to which bias is induced without changing the predictions outside the induced subset. In other words, instead of relearning the predictive model after differential sampling bias has been introduced into the subgroup, we directly calculate the new predictions from Theorem~\ref{thm:thm1}}. 
}

For each trial, we randomly partition the data into 80\% training and 20\% testing data.  If $\Delta > 1$, then differential sampling bias $\Delta$ is injected into subset $S^T$ for the training data $\widetilde D$, resampling data records $(\widetilde{x}_i,\widetilde{y}_i) \in \widetilde{D}_{S^T}$ with replacement (where records with $\widetilde{y}_i = 1$ have weight $\Delta$ and records with $\widetilde{y}_i = 0$ have weight 1), and leaving the test data $D$ and the rest of the training data unchanged. The classifier is trained on the biased training data, and used to make predictions $\widetilde p_i$ on the unbiased test data. Then Bias Scan is used to assess whether these predictions are biased, reporting the highest scoring subgroup $S^\ast = \arg\max_{S \in rect(X)} F(S)$ and its score $F^\ast=F(S^\ast)$.  We then compare the values of the Bias Scan score $F^\ast$, the score of the injected subgroup $F(S^T)$ (calculated by equation~(\ref{bias_scan_score})), and the theoretical score of $S^T$, which we denote as $F_{theo}(S^T)$.  The value of $F_{theo}(S^T)$ is computed using only the \emph{unbiased} training and test data, as defined in Theorem~\ref{thm:thm1}: $F_{theo}(S^T) = F_{old}(S^T) - \sum_{s_i \in D_{S^T}} y_i \log \Delta + \sum_{s_i \in D_{S^T}} \log (\Delta p_i + 1 - p_i)$, if $\Delta > \hat{q}_{MLE}$, and $F_{theo}(S^T) = 0$ otherwise.  We also compute the overlap (Jaccard coefficient) between the injected subset of test data records $D_{S^T}$ and the detected subset $D_{S^\ast}$:
\[ \text{overlap} = \frac{|D_{S_T} \cap D_{S^\ast}|}{|D_{S_T} \cup D_{S^\ast}|}.\]
Finally, we use Theorems~\ref{thm:thm2} and~\ref{thm:thm4} to estimate the critical value $h(\alpha)$ and the corresponding threshold value $\Delta_{thresh}$, for which we expect $\mathbb{P}(F^\ast > h(\alpha)) \rightarrow 1$ when $\Delta > \Delta_{thresh}$.  

Given these values for each amount of bias $\Delta$ (averaged over the 100 trials, for a given classifier and a given injected subgroup $S^T$), we form two plots: one comparing $F^\ast$, $F(S^T)$, and $F_{theo}(S^T)$ as a function of $\Delta$, and one showing overlap between $D_{S^T}$ and $D_{S^\ast}$ as a function of $\Delta$, as compared to $\Delta_{thresh}$.

If assumptions (A1)-(A4) hold, as the size of the training data grows to infinity, we expect perfect overlap between the curves for $F_{theo}(S^T)$ and $F(S^T)$ by Thm.~\ref{thm:thm1}. As $\Delta$ becomes large compared to $\Delta_{thresh}$, we expect $S^\ast \approx S^T$, and thus $F^\ast \approx F(S^T)$ and $\text{overlap} \approx 1$, while for small $\Delta$, we expect $F^\ast > F(S^T)$ and $\text{overlap} \ll 1$.  We now examine whether these expectations are met for the finite, real-world COMPAS dataset, for each classifier and each injected subgroup $S^T$.

\subsubsection{Experimental results}  
\begin{figure}[t]
  \centering
  a) \subfloat{\includegraphics[width=0.2\textwidth, height=0.2\textwidth] {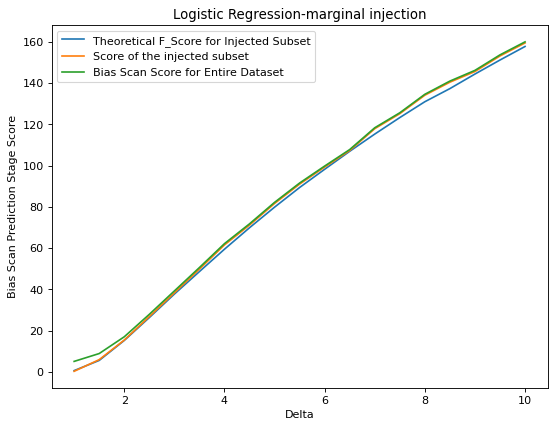}}
  b) \subfloat{\includegraphics[width=0.2\textwidth, height=0.2\textwidth] {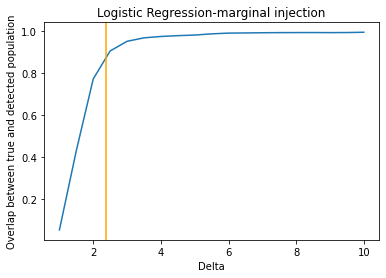}}
  \caption{Logistic regression classifier with marginal bias. (a) Scores $F^\ast$ (green), $F(S^T)$ (orange), and $F_{theo}(S^T)$ (blue) vs. $\Delta$. (b) Overlap vs. $\Delta$, as compared to $\Delta_{thresh}$.} \label{figure:fig2}
\end{figure}

\begin{figure}[t]
  \centering
  a) \subfloat{\includegraphics[width=0.2\textwidth, height=0.2\textwidth] {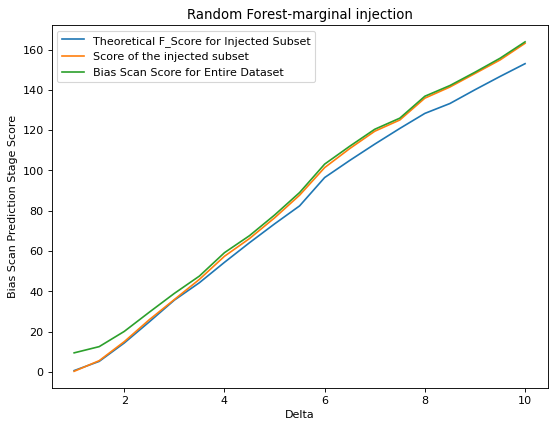}}
  b) \subfloat{\includegraphics[width=0.2\textwidth, height=0.2\textwidth] {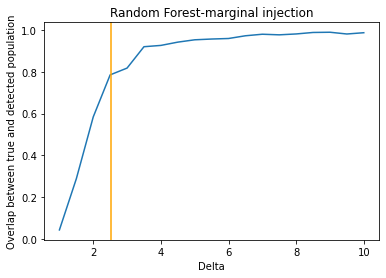}}
  \caption{Random forest classifier with marginal bias. (a) Scores $F^\ast$ (green), $F(S^T)$ (orange), and $F_{theo}(S^T)$ (blue) vs. $\Delta$. (b) Overlap vs. $\Delta$, as compared to $\Delta_{thresh}$.} \label{figure:fig3}
\end{figure}

\begin{figure}[t]
  \centering
  a) \subfloat{\includegraphics[width=0.2\textwidth, height=0.2\textwidth] {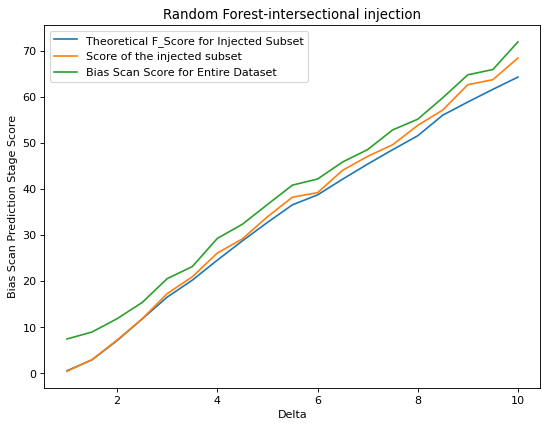}}
  b) \subfloat{\includegraphics[width=0.2\textwidth, height=0.2\textwidth] {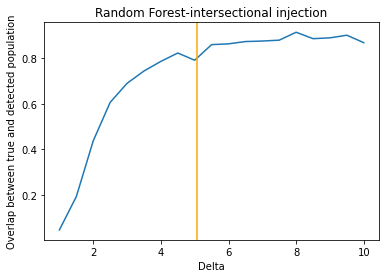}}
  \caption{Random forest classifier with intersectional bias. (a) Scores $F^\ast$ (green), $F(S^T)$ (orange), and $F_{theo}(S^T)$ (blue) vs. $\Delta$. (b) Overlap vs. $\Delta$, as compared to $\Delta_{thresh}$.} \label{figure:fig4}
\end{figure}

For the logistic regression classifier learned from training data injected with marginal differential sampling bias (Figure~\ref{figure:fig2}), we observe near-perfect overlap between the observed score $F(S^T)$ and theoretical score $F_{theo}(S^T)$ for the injected subgroup $S^T$, suggesting the validity of our theoretical results above.  As expected, the Bias Scan score $F^\ast \approx F(S^T)$ and $\text{overlap} \approx 1$ for $\Delta > \Delta_{thresh}$, while $F^\ast > F(S^T)$ and $\text{overlap} \ll 1$ for small $\Delta$.  For the random forest classifier learned from training data injected with marginal differential sampling bias (Figure~\ref{figure:fig3}), we see similar results, but with $F(S^T)$ slightly greater than $F_{theo}(S^T)$ for large $\Delta$.  This is likely due to data sparsity: the combination of finite training data and high bias may lead to few or no training data records with $\widetilde{y}_i = 0$ for some covariate profiles in the injected subgroup, leading to inaccurate estimation of $\widetilde{\mathbb{P}}(Y = 1 \:|\: X)$.  This pattern is repeated for the random forest classifier learned from training data injected with intersectional differential sampling bias (Figure~\ref{figure:fig4}), with a larger gap between $F(S^T)$ and $F_{theo}(S^T)$, most likely due to the smaller amount of training data in $S^T$.  Similarly, the smaller amount of test data in $S^T$ leads to some noise in the detected subgroup, resulting in $\text{overlap} \approx 0.9$ rather than 1, and thus $F^\ast = \max_{S\in rect(X)} F(S) > F(S^T)$.  Nevertheless, these results suggest that the theoretical values of $F_{theo}(S^T)$ and $\Delta_{thresh}$ are good approximations even for finite data.  


\begin{figure}[t]
  \centering
  a) \subfloat{\includegraphics[width=0.2\textwidth, height=0.2\textwidth] {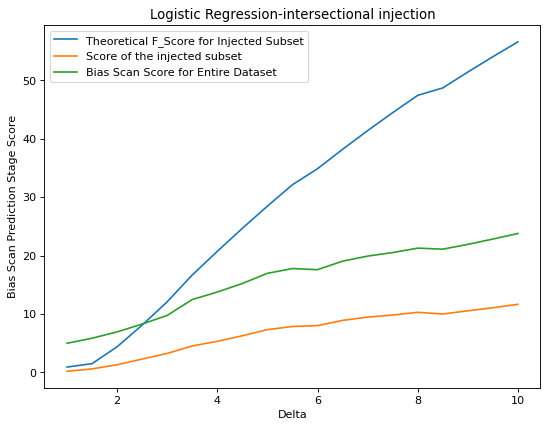}}
  b) \subfloat{\includegraphics[width=0.2\textwidth, height=0.2\textwidth] {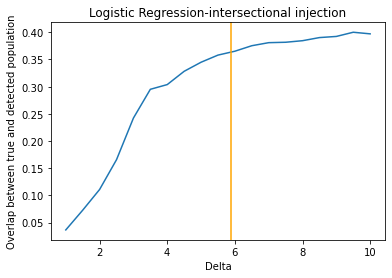}}
  \caption{Logistic regression classifier with intersectional bias. (a) Scores $F^\ast$ (green), $F(S^T)$ (orange), and $F_{theo}(S^T)$ (blue) vs. $\Delta$. (b) Overlap vs. $\Delta$, as compared to $\Delta_{thresh}$.} \label{figure:fig5}
\end{figure}

\begin{figure}[t]
  \centering
  a) \subfloat{\includegraphics[width=0.2\textwidth, height=0.2\textwidth] {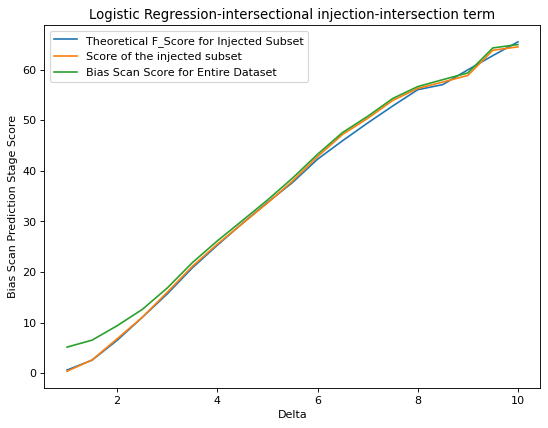}}
  b) \subfloat{\includegraphics[width=0.2\textwidth, height=0.2\textwidth] {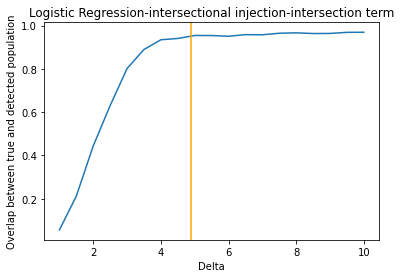}}
  \caption{Logistic regression classifier (including interaction term) with intersectional bias. (a) Scores $F^\ast$ (green), $F(S^T)$ (orange), and $F_{theo}(S^T)$ (blue) vs. $\Delta$. (b) Overlap vs. $\Delta$, as compared to $\Delta_{thresh}$.} \label{figure:fig6}
\end{figure}

For the logistic regression classifier learned from training data injected with intersectional differential sampling bias (Figure~\ref{figure:fig5}), however, we see a very different picture: as $\Delta$ increases, the Bias Scan score $F^\ast$ and the score of the injected subgroup $F(S^T)$ are both much smaller than the theoretical score $F_{theo}(S^T)$, and the overlap between $S^\ast$ and $S^T$ plateaus around 0.4 even for large $\Delta$. This is because assumption (A1) is violated: the logistic regression model is misspecified and cannot learn the intersectional bias against white females, instead learning separate (and much smaller) marginal biases against all females and all white individuals via the learned model coefficients on these terms.  When an interaction term for white females is manually added to the logistic regression model specification (Figure~\ref{figure:fig6}), we observe that this additional term resolves the problem, and we again have a near-perfect match between the theoretical and observed scores for the injected subgroup $S^T$.

\ignore{
\subsubsection{Experiments with Marginal Bias (Analysis I)}
\label{marginalbiasanalysis1}
\begin{enumerate}
    \item (Blue curve) is expected to differ from the (Yellow Curve) in practice, unlike the equality indicated in the score derivation in Eqs. \ref{scorederivation}, since the classifier is neither flexible nor trained upon large data in practice.
    \item (Green curve) is expected to be above the (Yellow Curve) as the Bias Scan score returns the maximum score across all rectangular subsets, which is greater than the score corresponding to a given subset $S$. (see Left Col. in Fig. \ref{analysis1}). , which is approximately equal to the theoretical score (see Left Col. in Fig. \ref{analysis1})
    \item The overlap coefficient increases as the amount of bias induced increases, and peaks to a maximum value of 1, because the classifier can identify the pattern that belonging to a sub-group is likely to produce the expected outcome. (see Right Col. in Fig. \ref{analysis2})
\end{enumerate}
}


\subsection{Experiments on NYPD Stop and Frisk Data}

The New York Police Department (NYPD) has long been plagued with accusations of racially discriminatory policing practices related to its ``stop, question, and frisk'' (SQF) policies. \citet{gelman07} found that persons of color ``were stopped more frequently than whites, even after controlling for precinct variability and race-specific estimates of crime participation''. \citet{goel2016precinct} concluded that Black and Hispanic individuals were disproportionately impacted by ``low hit rate'' stops, where the officer suspected the stopped individual of criminal possession of a weapon (CPW) but the \emph{ex ante} probability of recovering a weapon was low. Here we assess racial bias in NYPD policing practices by analyzing five years of SQF data during the peak of the stop and frisk policy, prior to a 2013 court ruling (Floyd v. City of New York) that NYPD stop-and-frisk tactics were unconstitutionally targeting New Yorkers of color.  

Thus our dataset consists of 760,489 pedestrian stops (made by NYPD officers for suspected CPW) from 2008-2012, downloaded from the city's web site\footnote{www1.nyc.gov/site/nypd/stats/reports-analysis/stopfrisk.page}. Following~\citet{goel2016precinct}, we first fit a logistic regression model to predict the probability that each stopped individual was found to have a weapon, using location (``housing'', ``transit'', or ``neither''), precinct, and 18 binary variables describing the circumstances of the stop\footnote{These circumstances include suspicious object, fits description, casing, acting as lookout, suspicious clothing, drug transaction, furtive movements, actions of violent crime, suspicious bulge, witness report, ongoing investigation, proximity to crime scene, evasive response, associating with criminals, changed direction, high crime area, time of day, and sights and sounds of criminal activity.} as predictors.  Stops with \emph{ex ante} probability of recovering a weapon at least 0.1 were marked as ``high probability''. If only high probability stops were conducted, 4.8\% of stops would have been made, 46\% of weapons would have been recovered, and the proportion of stopped individuals who were neither Black nor Hispanic would have more than doubled, from 9\% to 23\%.  

Next we create a new dataset with the demographics of each stopped individual (borough, sex, race, and age decile, all of which were excluded from the predictive model above), and whether each was a high or low probability stop.  We then assess racial bias by considering the race of the stopped individual as the outcome variable, and comparing the original, biased policing data to an alternative, ``less biased''\footnote{We refer to the high probability stop data as ``less biased'' rather than ``unbiased'' because it still contains biases based on which neighborhoods the NYPD officers chose to patrol, but eliminates the many low probability stops which predominantly and unfairly target racial minorities.} policing practice in which only high probability stops were made.  

More precisely, we perform the following steps, for each value of $k \in \{0,10,\ldots,100\}$: (1) split the data into equal-sized training and test sets; (2) remove all low probability stops from the test data; (3) remove ($100-k$)\% of the low probability stops from the training data; (4) learn a random forest classifier from the training data to estimate the probability that Race = Black for each stopped individual, conditional on the other demographic features; (5) use the learned model to predict the probability $\widetilde p_i$ that Race = Black for each stop in the test data; and (6) run Bias Scan on the predicted probabilities $\widetilde p_i$ and observed outcomes $y_i = \mathbf{1}\{\text{Race} = \text{Black}\}$ to identify the highest-scoring subgroup $S^\ast$ and its score $F^\ast = F(S^\ast)$.  Here $k=0$ corresponds to drawing the training data from the same, ``less biased'' distribution of stops as the test data, and $k=100$ corresponds to drawing the training data from the original, ``biased'' distribution of stops. 

\begin{figure}[t]
  \centering
  \includegraphics[width=0.2\textwidth, height=0.2\textwidth]{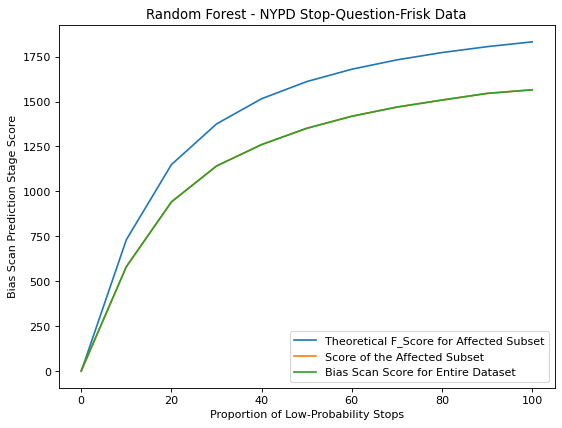} 
  \caption{Random forest classifier with heterogeneous bias (SQF data). Scores $F^\ast$ (green), $F(S^T)$ (orange), and $F_{theo}(S^T)$ (blue) vs. proportion of low probability stops $k$.}\label{figure:fig7}
\end{figure}

Thus, for $k > 0$, this process can be thought of as injecting differential sampling bias, increasing the odds that Race = Black by some factor $\Delta > 1$, as compared to the alternative policing practice of only making high probability stops.  However, this scenario poses several new challenges for our theoretical analysis: we do not know the injected subgroup $S^T$ or the amount of bias $\Delta$, and in fact the bias may be heterogeneous (different $\Delta$ for different covariate profiles).  Thus we make several simplifying assumptions. First, when auditing predictions from the model learned from the most biased training data ($k=100$), Bias Scan identifies a large, high-scoring subgroup $S^\ast$ consisting of individuals with $\text{Gender} \in \{\text{Male}, \text{Female}\}$, $\text{Age} < 70$, and $\text{Borough} \in \{\text{Manhattan}, \text{Brooklyn}, \text{Queens}, \text{Staten Island}\}$ (excluding the Bronx).  We assume that this $S^\ast$ is the true injected subgroup $S^T$.  Second, we assume that $\Delta$ is constant over subgroup $S^T$, and thus compute the odds ratio $\Delta = \frac{p_k(1-p_0)}{(1-p_k)p_0}$, where $p_k$ is the proportion of Black individuals in subgroup $S^T$ of the training dataset for a given value of $k$. Thus we have amounts of differential sampling bias ranging from $\Delta = 1$ for $k=0$ to $\Delta = 2.675$ for $k=100$. We then use these values of $\Delta$ along with the ``less biased'' training and test data ($k=0$) to plot $F_{theo}(S^T)$ as a function of $k$, and compare these theoretical values to the Bias Scan score $F^\ast$ and the subgroup score $F(S^T)$.  In Figure~\ref{figure:fig7}, we observe that $F^\ast = F(S^T)$ except when $k=0$, i.e., the same subgroup $S^\ast$ is detected for all $k>0$.  Additionally, we see that $F_{theo}(S^T)$ is a relatively good approximation for $F(S^T)$, with $F(S^T)$ consistently about 16\% lower than $F_{theo}(S^T)$ across all values of $k$.  This difference can be explained by our approximation of the heterogeneous bias $\Delta_x$, for covariate profiles $x \in S^T$, by estimating a single, constant $\Delta$ value.

\section{Conclusion}
It is critical both to analyze the downstream impacts of biases as they propagate through the learning pipeline, and to create new analytical tools to detect and mitigate propagating biases. With this work, we take a step toward these goals by quantifying how a particular data bias, differential sampling bias, propagates into biased model predictions, and providing theoretical guarantees for detection of the propagated biases. We validate our theoretical results through experiments on real-world criminal justice data where our assumptions are relaxed. In future work, we plan to extend our theoretical analysis of propagating biases to other types of data bias (e.g., measurement bias) as well as biases in other pipeline stages. We are particularly interested in analyzing when model predictions are impacted by multiple, interacting biases, which we believe is often the case in complex, real-world settings. 

\clearpage
\section{Acknowledgments}
This work was partially supported by the National Science Foundation Program on Fairness in Artificial Intelligence in Collaboration with Amazon, grant IIS-2040898. We gratefully acknowledge input from Prof. Ravi Shroff for designing experiments on the NYPD Stop and Frisk Data. 

\bibliography{aaai23}

\section{Technical Appendix}
\subsection{Derivation of Bias Scan Score $F(S)$}
\label{scorederivation}
To obtain the score function for a given subgroup $S$, Bias Scan computes the generalized log-likelihood ratio
$F(S) = \frac{\max_{\widetilde q} P(D \:|\: H_1(S,\widetilde q))}{P(D \:|\: H_0)}$, assuming the following hypotheses:
\begin{align*}
    &H_0 : &\text{odds}(y_i) = \frac{\widetilde{p}_i}{1-\widetilde{p}_i}, \:\: &\forall s_i \in D.~~~~~~~~ \\
    &H_1(S,\widetilde q) : &\text{odds}(y_i) = \frac{\widetilde{q} \: \widetilde{p}_i} {1-\widetilde{p}_i}, \:\: &\forall s_i \in D_S, \\ & &\text{odds}(y_i) = \frac{\widetilde{p}_i}{1-\widetilde{p}_i}, \:\: &\forall s_i \in D \setminus D_S.
\end{align*}

This implies that the probability $P(y_i) = \frac{\text{odds}(y_i)}{1+\text{odds}(y_i)} = \frac{\widetilde q \: \widetilde p_i}{1 - \widetilde p_i + \widetilde q \: \widetilde p_i}$ for $s_i \in D_S$ under $H_1(S,\widetilde q)$, and $P(y_i) = \widetilde p_i$ otherwise.  We denote these probabilities by $p_i^1$ and $p_i^0$ respectively, and derive the score $F(S)$:

\begin{align*}
    F(S) &= \max_{\widetilde{q}}
     \log \frac{\prod_{s_i \in D_S} {(p^1_i)}^{y_i} (1-p^1_i)^{1-y_i}}{\prod_{s_i \in D_S} {(p^0_i)}^{y_i}(1-p^0_i)^{1-y_i}} \\ &= \max_{\widetilde{q}}
     \left(\sum_{s_i \in D_S} y_i \log \frac{p^1_i}{p^0_i} + \sum_{s_i \in D_S} (1-y_i) \log \frac{1-p^1_i}{1-p^0_i}\right) \\ &= \max_{\widetilde{q}} \left( \sum_{s_i \in D_S} y_i \log \frac{\widetilde{q}}{1 - \widetilde{p}_i + \widetilde{q}\:\widetilde{p}_i} \right. \\  &~~~~~~~~~~~+ \left. \sum_{s_i \in D_S} (1-y_i) \log \frac{1}{1 - \widetilde{p}_i + \widetilde{q}\:\widetilde{p}_i} \right) \\ &= \max_{\widetilde{q}} \left( \sum_{s_i \in D_S} y_i \log \widetilde{q} - \sum_{s_i \in D_S} \log (1 - \widetilde{p}_i + \widetilde{q}\:\widetilde{p}_i) \right). 
\end{align*}

Here we focus on the case where the probabilities $\widetilde p_i$ are over-estimated, i.e., we identify biases where $\widetilde p_i > P(y_i)$ and thus $0 < \widetilde q < 1$.  Thus we define $F(S)$ as above for $\widetilde q_{MLE} = \arg\max_{\widetilde q} \left( \sum_{s_i \in D_S} y_i \log \widetilde{q} - \sum_{s_i \in D_S} \log (1 - \widetilde{p}_i + \widetilde{q}\:\widetilde{p}_i) \right) < 1$, and otherwise we have $\widetilde q_{MLE} = 1$ and thus $F(S) = 0$.

\subsection{Bias Scan Algorithm}
\label{biasscanmethod}

As described in~\citet{biasscan}, Bias Scan detects intersectional subgroups for which the classifier's probabilistic predictions $\widetilde p_i$ are significantly biased as compared to the observed binary outcomes $y_i$, by searching for the rectangular subgroup $S \in rect(X)$ which maximizes the log-likelihood ratio score $F(S)$ derived above.  

To optimize $F(S)$ over rectangular subgroups, Bias Scan performs a coordinate ascent procedure, optimizing $F(S)$ over subsets of values for one attribute at a time until convergence.  This coordinate ascent procedure is repeated for multiple iterations, starting from a different, randomly initialized rectangular subgroup on each iteration.  Bias Scan returns the maximum score $\hat F^\ast = \max_S F(S)$ and the corresponding subgroup $\hat S^\ast = \arg\max_S F(S)$ over all iterations. 

The Bias Scan algorithm consists of the following steps:

\begin{enumerate}
    \item Initialize $\hat F^\ast = 0$ and $\hat S^\ast = \emptyset$.
    \item Choose an initial rectangular subgroup $S = S_1 \times \ldots \times S_Q$ by randomly selecting a subset of values $S_j \subseteq V_j$, $S_j \ne \emptyset$, for each attribute $X_j$.  Mark all attributes as ``unvisited''.
    \item Randomly select an unvisited attribute $X_j$ and find $S_j' = \arg\max_{S_j \subseteq V_j, S_j \ne \emptyset} F(S_1 \times \ldots \times S_Q)$.  Let $S' = S_1 \times \ldots \times S_{j-1} \times S_j' \times S_{j+1} \times \ldots \times S_Q$.  
    \item If $F(S') > F(S)$, then set $S = S'$ and mark all attributes as ``unvisited''.
    \item If $F(S) > \hat F^\ast$, then set $\hat F^\ast = F(S)$ and $\hat S^\ast = S$.  
    \item Mark attribute $X_j$ as ``visited''.
    \item Repeat steps 3-6 until all attributes are marked as ``visited''.
    \item Repeat steps 2-7 for a fixed, large number of iterations $I$.
\end{enumerate}    

The optimization over subsets of values $S_j \subseteq V_j$ in Step 3 can be performed efficiently, requiring a number of computations of the score function which is linear rather than exponential in the arity $|V_j|$ of that attribute, thanks to the Linear Time Subset Scanning (LTSS) property of the Bias Scan score function~\cite{neill2012fast}.  The statistical significance of detected subgroups can be obtained by randomization testing, performing the same search procedure on a large number of datasets randomly generated under the null hypothesis $H_0$, and then comparing the score $\hat F^\ast$ for the original data to the $(1-\alpha)$ quantile of the distribution of $\hat F^\ast$ for the null data.  Further details are provided by~\citet{biasscan}.

\subsection{Proofs of Theorem 1 and Corollary 1}

\firstthm*
\begin{proof}
As $|\widetilde D| \rightarrow \infty$ without differential sampling bias, the number of training data records tends to infinity for each $x \in support(\widetilde f_X)$.  The classification model is consistent by assumption (A1), and thus the estimated probability $\hat p_i$ converges to $\mathbb{P}(Y = 1 | X = x) = p_i$ for all $x \in S$ for the training data.  By assumption (A2), $\hat p_i \rightarrow \mathbb{P}(Y = 1 | X = x) = p_i$ for all $x \in S$ for the test data, and the corresponding set of test data records $D_S$ is non-empty.  

Similarly, as $|\widetilde D| \rightarrow \infty$ with differential sampling bias $\Delta$ for subgroup $S$, the number of training data records tends to infinity for each $x \in support(\widetilde f_X)$. The classification model is consistent by assumption (A1), and thus the estimated probability $\widetilde p_i$ converges to $\widetilde{\mathbb{P}}(Y = 1 | X = x)$ = $\frac{\Delta\mathbb{P}(Y=1|X=x)}{\Delta\mathbb{P}(Y=1|X=x) + \mathbb{P}(Y=0|X=x)} = \frac{\Delta p_i}{\Delta p_i + 1 - p_i}$ for all $x \in S$ for the training data. By assumption (A2), $\widetilde p_i \rightarrow \widetilde{\mathbb{P}}(Y = 1 | X = x) = \frac{\Delta p_i}{\Delta p_i + 1 - p_i}$ for all $x \in S$ for the test data, and $D_S$ is non-empty. 

Next, we derive the relationship between the maximum likelihood estimate of the $\widetilde q$ parameter for Bias Scan, with and without differential sampling bias.  We define:
\[\widetilde q_{MLE} = 
\underset{\widetilde{q}}{\arg\max}
\left(
\sum_{s_i \in D_S} y_i \log \widetilde{q} - \sum_{s_i \in D_S} \log (1 - \widetilde{p}_i + \widetilde{q}\:\widetilde{p}_i)
\right),
\]
\[\hat q_{MLE} = \underset{\hat{q}}{\arg\max}
\left(
\sum_{s_i \in D_S} y_i \log \hat{q} - \sum_{s_i \in D_S} \log (1 - \hat{p}_i + \hat{q}\:\hat{p}_i)
\right).
\]
By setting $\frac{dF(S)}{d\widetilde{q}} = 0$ and $\frac{dF(S)}{d\hat{q}} = 0$ for the cases with and without differential sampling bias respectively, we obtain:
\[
    \sum_{s_i \in D_S} \frac{y_i}{\widetilde{q}_{MLE}} = \sum_{s_i \in D_S} \frac{\widetilde{p}_i}{1 - \widetilde{p}_i + \widetilde{q}_{MLE} \: \widetilde {p}_i},
\]
\[
    \sum_{s_i \in D_S} \frac{y_i}{\hat{q}_{MLE}} = \sum_{s_i \in D_S} 
    \frac{\hat{p}_i}{1 - \hat{p}_i + \hat{q}_{MLE} \: \hat {p}_i},
\]
and thus,
\[
\sum_{s_i \in D_S} \frac{\widetilde{q}_{MLE}\: \widetilde{p}_i}{1 - \widetilde{p}_i + \widetilde{q}_{MLE} \: \widetilde {p}_i} = \sum_{s_i \in D_S} \frac{\hat{q}_{MLE}\: \hat{p}_i}{1 - \hat{p}_i + \hat{q}_{MLE} \: \hat {p}_i}.
\]
Plugging in the values of $\hat p_i = p_i$ and $\widetilde p_i = \frac{\Delta p_i}{\Delta p_i + 1 - p_i}$ from above, and simplifying, we obtain:
\[
\sum_{s_i \in D_S} \frac{\widetilde{q}_{MLE}\: \Delta p_i}{1 - p_i + \widetilde{q}_{MLE} \: \Delta p_i} = \sum_{s_i \in D_S} \frac{\hat{q}_{MLE}\: p_i}{1 - p_i + \hat{q}_{MLE} \: p_i},
\]
and thus,
\[
    \widetilde{q}_{MLE}=\frac{\hat q_{MLE}}{\Delta}.
\]

We can now derive the Bias Scan score without differential sampling bias as:
\begin{align*}
F_{old}(S) &=  \sum_{s_i \in D_S} y_i \log \hat{q}_{MLE} - \sum_{s_i \in D_S} \log (1 - \hat{p}_i + \hat{q}_{MLE}\:\hat{p}_i) \\
&= 
\sum_{s_i \in D_S} y_i \log \hat{q}_{MLE} - \sum_{s_i \in D_S} \log (1 - p_i + \hat{q}_{MLE}\: p_i).
\end{align*}

Note that $F_{old}$ is defined without enforcing the constraint $\widetilde q < 1$.  Finally, we derive the Bias Scan score with differential sampling bias (for detecting over-estimated probabilities) as:
\begin{align*}
F(S) &= \sum_{s_i \in D_S} y_i \log \widetilde{q}_{MLE} - \sum_{s_i \in D_S} \log (1 - \widetilde{p}_i + \widetilde{q}_{MLE}\:\widetilde{p}_i) \\
&= \sum_{s_i \in D_S} y_i \log \frac{\hat q_{MLE}}{\Delta} 
- \sum_{s_i \in D_S} \log \left(1 +  \left(\frac{\hat q_{MLE}}{\Delta} - 1\right)\widetilde{p}_i\right) \\
&= \sum_{s_i \in D_S} y_i \log \frac{\hat q_{MLE}}{\Delta} 
- \sum_{s_i \in D_S} \log \left(1 + \frac{(\hat q_{MLE}-\Delta)p_i}{\Delta p_i + 1 - p_i}\right) \\ &= \sum_{s_i \in D_S} y_i \log \frac{\hat q_{MLE}}{\Delta} 
- \sum_{s_i \in D_S} \log \left(\frac{1 - p_i + \hat q_{MLE}p_i}{\Delta p_i + 1 - p_i}\right), 
\\&= F_{old}(S) - \sum_{s_i \in D_{S}} y_i \log \Delta + \sum_{s_i \in D_{S}} \log (\Delta p_i + 1 - p_i)
\end{align*}
if $\Delta > \hat q_{MLE}$ (and thus $\widetilde q_{MLE} = \hat q_{MLE}/\Delta < 1$), and otherwise we have 
$\widetilde q_{MLE} = 1$ and thus $F(S)=0$.  
\end{proof}

\firstcorr*
\begin{proof}
From Theorem~\ref{thm:thm1}, we have
\[ F(S) \rightarrow F_{old}(S) - \sum_{s_i \in D_{S}} y_i \log \Delta + \sum_{s_i \in D_{S}} \log (\Delta p_i + 1 - p_i), \]
if $\Delta > \hat q_{MLE}$. As $|D| \rightarrow \infty$, $\hat q_{MLE} \rightarrow 1$, and thus we have both w.h.p. $\Delta > \hat q_{MLE}$ and $F_{old}(S) \rightarrow 0$:
\[ F(S) \rightarrow 
\sum_{s_i \in D_S} (\log (\Delta p_i + 1 - p_i) - y_i \log \Delta), \]
and
\begin{align*}
\frac{F(S)}{|D|} \rightarrow \frac{|D_S|}{|D|} \mathbb{E}_{s_i \in D_S}[\log (\Delta p_i + 1 - p_i) - y_i \log \Delta].
\end{align*}
As $|D| \rightarrow \infty$, $|D_S|/|D| \rightarrow \mathbb{P}(x \in S)$, and 
$\mathbb{E}[y_i] = \mathbb{E}[\mathbb{E}[y_i\:|\:x_i]] = \mathbb{E}[p_i]$ for $s_i \in D_S$. Plugging in these values, we obtain the given expression.
\ignore
{
From Theorem~\ref{thm:thm1}, as $|D| \rightarrow \infty$, we have:
\[\frac{F(S)}{|D|} \rightarrow \frac{F_{old}(S)}{|D|} + \frac{|D_S|}{|D|} \mathbb{E}_{s_i \in D_S}[\log (\Delta p_i + 1 - p_i) - y_i \log \Delta]. \]
As $|D| \rightarrow \infty$, $|D_S|/|D| \rightarrow \mathbb{P}(x \in S)$.  Moreover, $\hat q_{MLE} \rightarrow 1$, and the ratio
$F_{old}(S)/|D| \rightarrow \mathbb{E}_{s_i \in D_S}[y_i \log \hat q_{MLE} - \log(1 - p_i + \hat q_{MLE}p_i)] = 0$.  Finally, as $|D_S| \rightarrow \infty$, $\mathbb{E}[y_i] = \mathbb{E}[p_i]$ for $s_i \in D_S$. Plugging in these values, we obtain the given expression.
}
To see that the expression increases with $\Delta$, assumption (A2) implies $\mathbb{P}(x \in S) > 0$, and the first derivative 
\[ \frac{d(\log (\Delta p_i + 1 - p_i) - p_i \log \Delta)}{d\Delta} = \frac{p_i}{\Delta p_i + 1 - p_i} - \frac{p_i}{\Delta} \]
is positive for $\Delta > 1$, given $0 < p_i < 1$ by assumption (A3).
\end{proof}

\subsection{Proof of Theorem 2 (and associated Lemmas)}
In this section, we derive a critical value $h(\alpha)$ of the Bias scan score $F^\ast$, for a given Type-I error rate $\alpha$, when no differential sampling bias is present.  Our approach is to upper bound $F^\ast = \max_{S\in rect(X)} F(S)$ by the Bias Scan score maximized over \emph{all} subgroups, $F_u^\ast = F(S_u^\ast)$, where $S_u^\ast = \arg\max_{S\subseteq V} F(S)$.  Additionally, as 
the number of training data records $|\widetilde D| \rightarrow \infty$, with no differential sampling bias, the classifier's predictions $\hat p_i$ converge to $p_i$ for all test records $s_i$, under assumptions (A1) and (A2), as in Theorem~\ref{thm:thm1}.  Thus we derive the distribution of $F_u^\ast$ under the null hypothesis, $H_0: \mathbb{P}(y_i = 1) = p_i$ for all $s_i \in D$.  

For any covariate profile $x$, we define the associated set of test data records $D_x = \{(x_i,y_i)\} \subseteq D: x_i = x$, and the aggregate quantities $y(x) = \sum_{s_i \in D_x} y_i$ and $n(x) = \sum_{s_i \in D_x} 1$.  We assume that the predicted probabilities $p_i$ are identical for all $s_i \in D_x$, since these data records have the same values for all predictor variables, and denote this probability by $p(x)$.  We can then write the Bias Scan score of a subgroup $F(S)$ as 
\[F(S) = \max_{0 < \widetilde{q} < 1} \sum_{x \in S}\gamma_x(\widetilde{q}),\] 
where
\[ \gamma_x(\widetilde{q}) = y(x) \log\left(\widetilde{q}\right) - n(x) \log (\widetilde{q} p(x) + 1 - p(x))\]
is the total contribution of data records with covariate profile $x$ to the score of subgroup $S$ for a given value of $\widetilde{q}$.  Then the maximum score over all subgroups can be written as
\begin{align*}
F_u^\ast &=  \max_{S \subseteq V} \max_{0 < \widetilde{q} < 1} \sum_{x \in S}\gamma_x(\widetilde{q}) \\
&= \max_{0 < \widetilde{q} < 1} \max_{S \subseteq V} \sum_{x \in S}\gamma_x(\widetilde{q}) \\
&= \max_{0 < \widetilde{q} < 1} \sum_{x \in V}\gamma_x(\widetilde{q}) \mathbf{1}\{\gamma_x(\widetilde{q}) > 0 \},    
\end{align*}
thus including all and only those covariate profiles which make a positive contribution to the score for the given value of $\widetilde{q}_{mle}^\ast$.
Given these definitions, we now consider the probability that a given covariate profile $x$ will have $\gamma_x(\widetilde{q}_{mle}^\ast) > 0$, and thus be included in $S_u^\ast$:

\begin{lemma}\label{lemma:lemma1}
Under the null hypothesis $H_0$, as $n(x) \rightarrow \infty$, the probability that covariate profile $x$ is included in the highest scoring subgroup $S_u^\ast$ converges to $1-\Phi\left(\frac{Z}{2}\sqrt{p(x)(1-p(x))}\right)$,
where $\widetilde{q}_{mle}^\ast = 1 - \frac{Z}{\sqrt{n(x)}}$, and $\Phi$ is the Gaussian cdf. 
\end{lemma}
\begin{proof}
Covariate profile $x$ is included in $S_u^\ast$ if and only if 
$\gamma_x(\widetilde{q}_{mle}^\ast) =  y(x) \log \left(\widetilde{q}_{mle}^\ast\right) - n(x) \log (\widetilde{q}_{mle}^\ast p(x) + 1 - p(x)) > 0$.  Given $0 < \widetilde{q}_{mle}^\ast < 1$, we have: 
\begin{align*}
&\mathbb{P}(\gamma_x(\widetilde{q}_{mle}^\ast) > 0) \\ &= \mathbb{P}\left( 
\frac{y(x)}{n(x)} < \frac{\log (\widetilde{q}_{mle}^\ast p(x) + 1 - p(x))}{\log \widetilde{q}_{mle}^\ast} \right) \\
&\rightarrow \mathbb{P}\left( 
\frac{y(x)}{n(x)} < p(x) - \frac{(1-\widetilde{q}_{mle}^\ast)p(x)(1 - p(x))}{2} \right) \\
&= \mathbb{P}\left( 
\psi(x) < - \frac{Z}{2} \sqrt{p(x)(1 - p(x))} \right),
\end{align*}
where $\psi(x) = \sqrt{\frac{n(x)}{p(x)(1-p(x))}}\left(\frac{y(x)}{n(x)} - p(x)\right)$. Here we have used a second order Taylor expansion for $\widetilde{q}_{mle}^\ast$, since $\widetilde{q}_{mle}^\ast$ converges to 1 as $n(x)\rightarrow\infty$. Next, since $y(x) \sim \text{Binomial}(n(x),p(x))$ under $H_0$, $\psi(x) \rightarrow\text{Gaussian}(0,1)$ as $n(x)\rightarrow\infty$, 
and $\mathbb{P}(\gamma_x(\widetilde{q}_{mle}^\ast) > 0)$ converges to
$1- \Phi\left(\frac{Z}{2}\sqrt{p(x)(1-p(x))}\right)$, where $\Phi$ is the Gaussian cdf.
\end{proof}

\begin{lemma}\label{lemma:lemma2}
Under the null hypothesis $H_0$, as $n(x) \rightarrow \infty$, the 
expectation and variance of $\gamma_x(\widetilde{q}_{mle}^\ast) \mathbf{1}\{\gamma_x(\widetilde{q}_{mle}^\ast)>0\}$ are upper bounded by constants $k_1 \approx 0.202$ and $k_2^2 \approx 0.274$ respectively.
\end{lemma}
\begin{proof}
From Lemma~\ref{lemma:lemma1}, as $n(x)\rightarrow\infty$, $\psi(x) = \sqrt{\frac{n(x)}{p(x)(1-p(x))}}\left(\frac{y(x)}{n(x)}-p(x)\right) \rightarrow \text{Gaussian}(0,1)$.  Moreover, conditional on $\gamma_x(\widetilde{q}_{mle}^\ast) > 0$, $\psi(x)$ has its right tail truncated at $\beta = -\frac{Z}{2}\sqrt{p(x)(1-p(x))}$, giving $\mathbb{E}[\psi(x) \:|\: \gamma_x(\widetilde{q}_{mle}^\ast) > 0] = -h(-\beta)$, and $\text{Var}[\psi(x) \:|\: \gamma_x(\widetilde{q}_{mle}^\ast) > 0] = 1 - \beta h(-\beta) - h(-\beta)^2$, where $h(x)=\frac{\phi(x)}{1-\Phi(x)}$ is the Gaussian hazard function.  
Since $y(x) = n(x)p(x) + \psi(x)\sqrt{n(x)p(x)(1-p(x))}$, 
this implies:
\begin{align*}
    &\mathbb{E}\left[\frac{y(x)-n(x)p(x)}{\sqrt{n(x)}} \:|\: \gamma_x(\widetilde{q}_{mle}^\ast) > 0\right] \nonumber \\ &= -h(-\beta)\sqrt{p(x)(1-p(x))}; \\ &\text{Var}\left[\frac{y(x)-n(x)p(x)}{\sqrt{n(x)}} \:|\:\gamma_x(\widetilde{q}_{mle}^\ast) > 0\right] \nonumber \\ &= (1 - \beta h(-\beta) - h(-\beta)^2)p(x)(1-p(x)).
\end{align*}
Next, as in Lemma~\ref{lemma:lemma1}, we can use a second-order Taylor expansion to write:
\begin{align*}
&\gamma_x(\widetilde{q}_{mle}^\ast) \\
&=  y(x) \log \widetilde{q}_{mle}^\ast - n(x) \log (\widetilde{q}_{mle}^\ast p(x) + 1 - p(x)) \\
&=  y(x) \log \widetilde{q}_{mle}^\ast - n(x) \log \widetilde{q}_{mle}^\ast \frac{\log (\widetilde{q}_{mle}^\ast p(x) + 1 - p(x))}{\log \widetilde{q}_{mle}^\ast} \\ &\rightarrow  (y(x)-n(x)p(x)) \log \widetilde{q}_{mle}^\ast \\ &~~~+ \frac{(1-\widetilde{q}_{mle}^\ast)(\log \widetilde{q}_{mle}^\ast)n(x)p(x)(1-p(x))}{2} \\
&\rightarrow  (y(x)-n(x)p(x)) \left( -\frac{Z}{\sqrt{n(x)}} \right) ~ \\ &~~~+ \frac{1}{2}\left(\frac{Z}{\sqrt{n(x)}} \right)\left( -\frac{Z}{\sqrt{n(x)}} \right)n(x)p(x)(1-p(x)) \\
&= -Z \left( \frac{y(x)-n(x)p(x)}{\sqrt{n(x)}} \right) - \frac{Z^2}{2} p(x)(1-p(x)),
\end{align*}
where we have used $\log \widetilde{q}_{mle}^\ast \rightarrow -(1-\widetilde{q}_{mle}^\ast) = -\frac{Z}{\sqrt{n(x)}}$ as $n(x)\rightarrow\infty$.

Then for the expectation we have:
\begin{align*}
&\mathbb{E}[\gamma_x(\widetilde{q}_{mle}^\ast) \:|\: \gamma_x(\widetilde{q}_{mle}^\ast) > 0] \\
&\rightarrow -Z \mathbb{E}\left[ \frac{y(x)-n(x)p(x)}{\sqrt{n(x)}} \right] - \frac{Z^2}{2} p(x)(1-p(x)) \\
&= Zh(-\beta)\sqrt{p(x)(1-p(x))} - \frac{Z^2}{2} p(x)(1-p(x)) \\
&= -2\beta h(-\beta) - 2\beta^2.
\end{align*}
Then, since $\mathbb{P}(\gamma_x(\widetilde{q}_{mle}^\ast) > 0) = 1-\Phi(-\beta)$ from Lemma~\ref{lemma:lemma1}, we know:
\begin{align*}
&\mathbb{E}[\gamma_x(\widetilde{q}_{mle}^\ast) \mathbf{1}\{ \gamma_x(\widetilde{q}_{mle}^\ast) > 0\}] \\ &\rightarrow (1-\Phi(-\beta))(-2\beta h(-\beta) - 2\beta^2)\\ 
&= -2\beta\phi(-\beta) - 2\beta^2(1-\Phi(-\beta))\\
&\le k_1,
\end{align*}
since this expression attains a maximum value of $k_1 \approx 0.202456$ at $\beta \approx -0.61$. 

\noindent For the variance, we have:
\begin{align*}
&\text{Var}[\gamma_x(\widetilde{q}_{mle}^\ast) \:|\: \gamma_x(\widetilde{q}_{mle}^\ast) > 0] \\
&\rightarrow Z^2 \text{Var}\left[ \frac{y(x)-n(x)p(x)}{\sqrt{n(x)}} \right] \\
&= Z^2 (1 - \beta h(-\beta) - h(-\beta)^2)p(x)(1-p(x)) \\
&= 4\beta^2 (1 - \beta h(-\beta) - h(-\beta)^2).
\end{align*}
\noindent Then, we know:
\begin{align*}
&\text{Var}[\gamma_x(\widetilde{q}_{mle}^\ast) \mathbf{1}\{ \gamma_x(\widetilde{q}_{mle}^\ast) > 0\}] \\ &= \text{Var}[\gamma_x(\widetilde{q}_{mle}^\ast) \:|\: \gamma_x(\widetilde{q}_{mle}^\ast) > 0] \mathbb{P}(\gamma_x(\widetilde{q}_{mle}^\ast) > 0) ~+ \\
&~~~~~ \mathbb{E}[\gamma_x(\widetilde{q}_{mle}^\ast) \:|\: \gamma_x(\widetilde{q}_{mle}^\ast) > 0]^2
\mathbb{P}(\gamma_x(\widetilde{q}_{mle}^\ast) > 0)\\
&~~~~~ (1-\mathbb{P}(\gamma_x(\widetilde{q}_{mle}^\ast) > 0)) \\
&\rightarrow 4\beta^2 (1 - \beta h(-\beta) - h(-\beta)^2)(1-\Phi(-\beta)) ~+ \\ &~~~~ (-2\beta h(-\beta) - 2\beta^2)^2(1-\Phi(-\beta))\Phi(-\beta) \\
&= 4\beta^2 (1-\Phi(-\beta)) (1 - \beta h(-\beta) - h(-\beta)^2 + \\ &~~~~ (\beta+h(-\beta))^2 \Phi(-\beta)) \\
&\le k_2^2,
\end{align*}
since this expression attains a maximum value of $k_2^2 \approx 0.273709$ at $\beta \approx -0.98$. \\
\end{proof}

\secondthm*
\begin{proof}
As $|\widetilde D| \rightarrow \infty$ without differential sampling bias, the number of training data records tends to infinity for each $x \in support(\widetilde f_X)$.  The classification model is consistent by assumption (A1), and thus the estimated probability $\hat p_i$ converges to $\mathbb{P}(Y = 1 | X = x) = p_i$ for all $x \in S$ for the training data.  By assumption (A2), $\hat p_i \rightarrow \mathbb{P}(Y = 1 | X = x) = p_i$ for all $x \in S$ for the test data, and the corresponding set of test data records $D_S$ is non-empty. As shown above, $F^\ast \le F_u^\ast = \sum_{x \in V} \gamma_x(\widetilde{q}_{mle}^\ast) \mathbf{1}\{ \gamma_x(\widetilde{q}_{mle}^\ast) > 0 \}$,
where  $\widetilde{q}_{mle}^\ast = \arg\max_{0<\widetilde{q}<1} \sum_{x\in V} \gamma_x(\widetilde{q}) \mathbf{1}\{ \gamma_x(\widetilde{q}) > 0\}$ and $\gamma_x(\widetilde{q}) =  y(x) \log \widetilde{q} - n(x) \log (\widetilde{q} p(x) + 1 - p(x))$.  From Lemma~\ref{lemma:lemma2}, for each of the $M$ unique covariate profiles $x$ in the test data, we know that $\gamma_x(\widetilde{q}_{mle}^\ast) \mathbf{1}\{ \gamma_x(\widetilde{q}_{mle}^\ast) > 0\}$
is drawn from a censored Gaussian distribution, with mean $\mu_x \le k_1$ and variance $\sigma_x^2 \le k_2^2$, where $k_1 \approx 0.202$ and $k_2 \approx \sqrt{0.274} \approx 0.523$.  Moreover, since a censored Gaussian with bounded variance has bounded fourth moment, we know that the Lyapunov condition holds.  Thus, from the Lyapunov CLT, we know that for large $M$, $\frac{F_u^\ast - \sum_{x \in V}\mu_x}{\sqrt {\sum_{x \in V} \sigma_x^2}} \rightarrow \text{Gaussian}(0,1)$, and by assumption (4) we know that $M$ is large enough for $F_u^\ast$ to be approximately Gaussian.  Then since $F^\ast \le F_u^\ast$, $\mu_x \le k_1 \:\forall x$, and $\sigma_x^2 \le k_2^2 \:\forall x$, we have:
\begin{align*}
&\mathbb{P}(F^\ast > h(\alpha)) \\
&= \mathbb{P}(F^\ast > k_1 M + k_2 \Phi^{-1}(1-\alpha) \sqrt{M}) \\
&\le \mathbb{P}(F_u^\ast > k_1 M + k_2 \Phi^{-1}(1-\alpha) \sqrt{M}) \\
&= \mathbb{P}\left(F_u^\ast > \left(\sum_{x\in V} k_1 \right) + \Phi^{-1}(1-\alpha) \sqrt{\sum_{x\in V} k_2^2}\right) \\
&\le \mathbb{P}\left(F_u^\ast > \left( \sum_{x\in V} \mu_x \right) + \Phi^{-1}(1-\alpha) \sqrt{\sum_{x\in V} \sigma_x^2}\right) \\
&= 1-\Phi(\Phi^{-1}(1-\alpha)) \\
&= \alpha.
\end{align*}
\end{proof}

\subsection{Proofs of Theorems 3 and 4}
\thirdthm*
\begin{proof}
By Corollary~\ref{corr:corr1}, as $|D| \rightarrow \infty$, $F(S^T)/|D|$ converges to $\mathbb{P}(x \in S^T) \mathbb{E}_{s_i \in D_{S^T}} [\log (\Delta p_i + 1 - p_i) - p_i \log \Delta]$, which is greater than zero because $\mathbb{P}(x\in S^T) > 0$ by assumption (A2), $0 < p_i < 1$ by assumption (A3), and $\log (\Delta p_i + 1 - p_i) - p_i \log \Delta > 0$ when $\Delta > 1$ and $0 < p_i < 1$.  By Theorem~\ref{thm:thm2}, as $|D| \rightarrow \infty$, for any $\alpha > 0$, $h(\alpha)$ converges to a constant independent of $|D|$.  Thus $h(\alpha)/|D| \rightarrow 0$ and 
$\mathbb{P}(F(S^T) > h(\alpha)) \rightarrow 1$.  Finally, since subgroup $S^T$ is rectangular, $F^\ast = \max_{S \in rect(X)} F(S) \ge F(S^T)$,
and $\mathbb{P}(F^\ast > h(\alpha)) \rightarrow 1$.
\end{proof}

\fourththm*
\begin{proof}
From Theorem~\ref{thm:thm1}, for finite $|D_{S^T}|$, we have $F(S^T) \rightarrow F_{old}(S^T) + Q(\Delta)$ for $\Delta > \hat q_{MLE}$ and $|\widetilde D| \rightarrow \infty$. We derive:
\begin{align*}
\frac{dQ}{d\Delta} &= 
\sum_{s_i \in D_{S^T}} \left(\frac{p_i}{\Delta p_i + 1 - p_i} - \frac{y_i}{\Delta}\right) \\
&= \frac{1}{\Delta} \sum_{s_i \in D_{S^T}} (\widetilde p_i - y_i). 
\end{align*}

Since $0 < p_i < 1$ by assumption (A3), all $\widetilde p_i$ are increasing with $\Delta$.  Moreover, since $\sum_{s_i \in D_{S^T}} (\widetilde p_i - y_i) = 0$ for $\Delta = \hat q_{MLE}$, $\sum_{s_i \in D_{S^T}} (\widetilde p_i - y_i) > 0$ for all $\Delta > \hat q_{MLE}$.  This implies that $Q(\Delta)$ is increasing, and therefore invertible, on the interval $\Delta \ge \hat q_{MLE}$.

Next we show $Q(\Delta) \rightarrow \infty$ as $\Delta \rightarrow \infty$. For some small positive $\epsilon \approx 0$, let $\Delta_\epsilon$ denote the minimum value of $\Delta > \hat q_{MLE}$ such that $\sum_{s_i \in D_{S^T}} (\widetilde p_i - y_i) \ge \epsilon$.  Then
for any $\Delta' > \Delta_\epsilon$, we have:
\begin{align*}
Q(\Delta') &= Q(\Delta_\epsilon) + \int_{\Delta_\epsilon}^{\Delta'} \frac{dQ}{d\Delta} \: d\Delta \\
&\ge Q(\Delta_\epsilon) + \int_{\Delta_\epsilon}^{\Delta'} \frac{\epsilon}{\Delta} \: d\Delta \\
&= Q(\Delta_\epsilon) + \epsilon(\log \Delta' - \log \Delta_\epsilon) \\
&= C_1 \log \Delta' + C_0
\end{align*}
for constants $C_1$ and $C_0$, and thus $Q(\Delta)$ increases as $o(\log \Delta)$ for $\Delta \ge \hat q_{MLE}$.  

Now, since $F_{old}(S^T)$ is independent of $\Delta$, we know that $F_{old}(S^T) + Q(\Delta)$ is continuous and increasing for $\Delta \ge \hat q_{MLE}$, and $\lim_{\Delta\rightarrow\infty} F_{old}(S^T) + Q(\Delta) = \infty$.  Since 
$F_{old}(S^T) + Q(\Delta) = 0$ at $\Delta = \hat q_{MLE}$, there must exist a single intermediate value of $\Delta > \hat q_{MLE}$ such that $F_{old}(S^T) + Q(\Delta) = h(\alpha)$, i.e., $\Delta = Q^{-1}(h(\alpha)-F_{old}(S^T))$.  Then we set $\Delta_{thresh} = \max(1, Q^{-1}(h(\alpha)-F_{old}(S^T)))$. This implies that $F(S^T) \rightarrow F_{old}(S^T) + Q(\Delta) > h(\alpha)$, and $\mathbb{P}(F(S^T) > h(\alpha)) \rightarrow 1$, for $\Delta > \Delta_{thresh}$.
Finally, assuming that subgroup $S^T$ is rectangular, $F^\ast = \max_{S \in rect(X)} F(S) \ge F(S^T)$, and $\mathbb{P}(F^\ast > h(\alpha)) \rightarrow 1$ for $\Delta > \Delta_{thresh}$.
\end{proof}

\end{document}